\documentclass{article}
\usepackage{cite}
\usepackage{spconf}
\usepackage{amsmath,amssymb,amsfonts, amscd, amsthm, dsfont, amsopn}
\usepackage{graphicx}
\usepackage{textcomp}
\usepackage{xcolor}
\usepackage{array}
\usepackage{amsthm}
\usepackage{subfigure,enumitem}
\usepackage{comment,hyperref}
\usepackage{algorithm}
\usepackage{algpseudocode}
\usepackage{booktabs}

\def\BibTeX{{\rm B\kern-.05em{\sc i\kern-.025em b}\kern-.08em
    T\kern-.1667em\lower.7ex\hbox{E}\kern-.125emX}}
\usepackage{color}
\usepackage{xcolor}

\allowdisplaybreaks

\DeclareMathOperator*{\argmin}{arg\,min}

\title{Augment on Manifold: Mixup Regularization with UMAP}

\name{Yousef El-Laham$^\star$ \qquad Elizabeth Fons$^\star$ \qquad Dillon Daudert$^\dagger$\qquad Svitlana Vyetrenko$^\star$}
\address{J.P. Morgan AI Research$^\star$ \qquad Stony Brook University$^\dagger$}
\usepackage{setspace}  

\begin{document}
\ninept

\maketitle

\begin{abstract}
Data augmentation techniques play an important role in enhancing the performance of deep learning models. Despite their proven benefits in computer vision tasks, their application in the other domains remains limited. This paper proposes a Mixup regularization scheme, referred to as UMAP Mixup, designed for ``on-manifold" automated data augmentation for deep learning predictive models. The proposed approach ensures that the Mixup operations result in synthesized samples that lie on the data manifold of the features and labels by utilizing a dimensionality reduction technique known as uniform manifold approximation and projection. Evaluations across diverse regression tasks show that UMAP Mixup is competitive with or outperforms other Mixup variants, show promise for  its potential as an effective tool for enhancing the generalization performance of deep learning models. 
\end{abstract}

\begin{keywords}
data augmentation, Mixup regularization, dimensionality reduction, UMAP
\end{keywords}

\section{Introduction}
Data augmentation (DA) has been extensively explored for deep learning (DL) models in computer vision (see \cite{shorten2019survey} for a review). Mainly, DA plays a pivotal role in improving the generalization of DL models, especially in contexts where available training data is scarce or imbalanced, whereby DL models are more susceptible to overfitting. Fundamentally, DA helps to implictly regularize a DL model to capture underlying invariances in the data, which is an essential aspect of model generalization. As an example, the label of an image remains invariant across a variety of transformations, such as rotations, cropping, grayscaling, translation, and flipping. These transformations effectively expand the diversity of the training data, which in turn boosts the model's ability to generalize and improves its performance on out-of-sample data. Despite these advances, the application of DA for DL models in other data modalities, such as tabular data or time series data, remains less explored. As an example, financial time series data poses unique challenges that standard data augmentation techniques might not adequately address. While there have been exploratory studies focusing on data augmentation for financial time series \cite{fons2020evaluating}, a generalized framework that can affirm the benefits of synthetic data augmentation across diverse financial datasets is yet to be established. This challenge derives from the fact that invariance properties of financial data are not as obvious as those in computer vision. 

In recent years, a technique known as Mixup has gained traction as a domain-agnostic form of data augmentation \cite{zhang2017mixup}. Mixup works by generating synthetic samples as convex combinations of two random samples during the training process. 
Notably, Mixup has shown a significant boost in performance comparison to standard training, especially on larger computer vision datasets. At its core, Mixup can be seen as a vicinal risk minimization (VRM) technique that aims to regularize the learning algorithm by encouraging it to behave linearly in-between observed samples \cite{chapelle2000vicinal}. This stands in contrast to empirical risk minimization (ERM), a principle of statistical learning theory that focuses on minimizing the loss over the empirical distribution of the training data \cite{vapnik1999overview}. Over the past several years, a variety of Mixup variants have been proposed, each addressing specific challenges or application domains. These include Manifold Mixup \cite{verma2019manifold}, Remix \cite{chou2020remix}, local Mixup \cite{baena2022preventing} and C-Mixup \cite{yao2022c}, among others. For instance, Manifold Mixup extends the concept of Mixup to the hidden layers of a neural network, applying Mixup in an embedding obtained at some intermediate layer of the network, thereby promoting smoother behavior in the overall model. Another variant, C-Mixup, was introduced  for regression tasks, and demonstrates some theoretical guarantees regarding its predictive performance over the standard Mixup technique for certain and simple classes of predictive models. Despite these advances, Mixup and its variants face several challenges. It remains difficult to derive a comprehensive theoretical framework that guarantees that Mixup will consistently improve the performance on out-of-sample, especially in contexts where the data may be subject to distribution shifts. Furthermore, for each Mixup technique, there is no way to guarantee that the mixing procedure of the features and labels will yield a synthesized sample that lies on the data manifold \cite{guo2019mixup}. While the Mixup technique has been touted as a domain-agnostic method, its empirical evaluation and validation, particularly in the context of data modalities other than computer vision, remain sparse. 

This work presents a novel Mixup regularization scheme for deep learning models that addresses some of these challenges. Specifically, we introduce a variant of Manifold Mixup that aims to guarantee that the Mixup operation results in a synthesized sample that lies on the data manifold. This is accomplished by training an intermediate layer of the predictor to learn an embedding that exhibits this property, using a technique called uniform manifold approximation and projection (UMAP) \cite{mcinnes2018umap}. UMAP is a nonlinear dimensionality reduction method that optimizes embeddings to preserve the topological structure of the data. The proposed approach, which we refer to as UMAP Mixup, incorporates an additional UMAP-based regularizer during training to optimize the learned embedding used for constructing the Mixup augmentations. The UMAP-based regularizer encourages the model to assign similar latent embeddings to the data points that are near each other in feature space according to a pre-defined distance metric, thereby promoting better on-manifold data augmentation. Empirically, we demonstrate that UMAP Mixup obtains competitive or better performance than ERM and a variety of Mixup variants on various regression tasks. 


\section{Problem Formulation}
\label{sec:formulation}
This work studies  DA in supervised learning problems from the perspective of statistical learning theory.  In supervised learning, we are interested in finding a function $f\in{\cal F}$ that describes the relationship between a feature vector $x\in\mathbb{R}^{d_x}$ and an output $y\in\mathbb{ R}^{d_y}$. 
Suppose that $x$ and $y$  have a joint cumulative distribution function $F(x, y)$. Our goal is in finding a function $f\in{\cal F}$ that minimizes the risk:
\begin{equation}
    \label{eq: expected_risk}
    R[f] = \mathbb{E}[\ell(f(x), y)] = \int \ell(f(x), y)dF(x, y),
\end{equation}
where $\ell$ denotes a loss function used to measure the discrepancy between function output and the true output. In practice,  it is common to assume $f$ belongs to some parametric family of functions ${\cal F}_{\theta}$ defined by parameters $\theta\in\Theta$. In this context, we would like to solve the following optimization problem:
\begin{equation}
    \label{eq: minimizing_expected_risk}
    \widehat\theta = \argmin _{\theta \in \Theta} R(\theta), 
\end{equation}
where $R(\theta)=\mathbb{E}[\ell(f_\theta(x), y)]$. Unfortunately, $F(x, y)$ is unknown and we must resort to using an approximations of the risk in order to solve \eqref{eq: minimizing_expected_risk}. Typically, the ERM priniciple is utilized, whereby parameters are learned by minimizing an approximation of the risk using the empirical distribution of an observed dataset ${\cal D}=\{(x_i, y_i)\}_{i=1}^N$ :
\begin{equation}
    \label{eq: erm}
    \widehat\theta_{\rm ERM} = \argmin_{\theta\in\Theta} \frac{1}{N}
    \sum_{i=1}^N \ell(f_{\theta}(x_i), y_i).
\end{equation}
The ERM principle results in a good parameter estimate if ${\cal D}$ accurately captures the true distribution $F(x, y)$. This is not typically true in practice, and so regularization techniques are often required to help predictive models generalize better to out-of-sample data.


\section{Background}

\subsection{Mixup Regularization}
Mixup is a VRM technique introduced to improve the generalization performance of deep learning models. The central idea of Mixup is to train with random convex combinations of training samples, rather than the original samples themselves. Mixup augmentations are constructed as follows:
\begin{align}
    \tilde{x} &= \lambda x_{i} + (1-\lambda) x_{j} \\
    \tilde{y} &= \lambda y_{i} + (1-\lambda) y_{j},
\end{align}
where $(x_i, y_i)$ and $(x_j, y_j)$ are independently sampled from the empirical distribution of the observed dataset, $\lambda\sim {\rm Beta}(\alpha, \alpha)$, and $\alpha$ is a hyperparameter controlling the mixing of the samples. Empirically, Mixup has been shown to improve the test performance of deep learning models. Recent theoreotical works attribute this improvement in performance to the fact Mixup is a form of implicit regularization \cite{zou2023mixupe}. An important extension in Mixup, called Manifold Mixup, performs the Mixup operation in an intermediate layer of the deep learning model and in some cases can be shown to improve on the performance of the original Mixup algorithm.  Unfortunately, Manifold Mixup does not guarantee that augmentations are synthesized on a manifold with nice properties (e.g, preservation of both local and global structure). 



\subsection{UMAP}
UMAP is a nonlinear dimensionality reduction technique that aims to learn a lower-dimensional embedding of the data that preserves its topological structure. A strong advantage of UMAP versus other dimensionality reduction techniques, such as T-SNE \cite{van2008visualizing}, is its ability to capture both local and global structure in the learned embeddings. 

UMAP makes three assumptions of the data: the data is assumed to be uniformly distributed on a Riemannian manifold; the Riemmanian metric is assumed to be locally constant; and the manifold is assumed to be locally connected. Based on these assumptions, one can optimize a set of embeddings  $\{z_i\in\mathbb{R}^{d_z}\}_{i=1}^N$, where each $z_i$ corresponds to the embedding of the data point $x_i$. 
The basic procedure has two steps: First, a topological representation of the data is constructed using fuzzy simplicial sets, and second, the embeddings are optimized so that their topological representation has the closest fuzzy topological structure to that of the representation in data space. 
Computationally, this amounts to minimizing the cross entropy between two graph representations: $P$ (data graph) and $Q$ (embedding graph). Nonparametric UMAP optimizes the embeddings directly, whereas parametric UMAP optimizes the parameters of some function approximator (e.g., a neural network) that outputs the embeddings \cite{sainburg2021parametric}. Since, this work focuses on embeddings obtained from neural networks, for brevity, we only highlight the parametric UMAP procedure.

The data graph $P$, whose elements are denoted by $p_{i, j}$, is constructed by computing directional probabilities between each datapoint and its $K$-nearest neighbors given by:
\begin{equation}
    p_{j|i} = \exp\left(\frac{-(d(x_i, x_j)-\rho_i)}{\sigma_i}\right), \quad x_j\in{\cal N}(x_i)
\end{equation}
where $d(x_i, x_j)$ is a distance function, $\rho_i$ and $\sigma_i$ are local connectivity parameters, and ${\cal N}(x_i)$ is the local neighborhood of $x_i$. The local connectivity parameters and the local neighborhood are  determined by the $K$ number of neighbors considered, which is a hyperparameter in the UMAP algorithm. Once the directional probabilities $p_{j|i}$ are determined, the probabilities are symmetrized
\begin{equation}
    p_{i, j} = (p_{j|i}+p_{i|j}) - p_{j|i}p_{i|j}
\end{equation}
Let $z_i=h_{\phi}(x_i)$ denote the embedding of data point $x_i\in\mathbb{R}^{d_x}$ obtain from some parametric function $h_{\phi}:\mathbb{R}^{d_x}\rightarrow \mathbb{R}^{d_z}$ with $\phi\in\Phi$. The embedding graph $Q_{\phi}$, whose elements edges are denoted by $q_{i, j}$, is constructed by computing:
\begin{equation}
    q_{i, j} = (1+ a\|z_i-z_j\|^{2b})^{-1},
\end{equation}
where $a$ and $b$ are hyperparameters that impact the minimum distance between the embeddings. Given the data graph $P$, the UMAP algorithm optimizes the embeddings such that the corresponding embedding graph $Q_{\phi}$ is the one that minimizes the cross-entropy loss between $P$ and $Q_{\phi}$, which is defined as 
\begin{equation}
    \label{eq: umap_loss}
    C(P, Q_{\phi}) = \sum_{i\neq j} p_{i,j} \log \left(\frac{p_{i,j}}{q_{i, j}}\right) + (1-p_{i,j})\log \left(\frac{1-p_{i,j}}{1-q_{i, j}}\right)
\end{equation}
In parametric UMAP, optimization of \eqref{eq: umap_loss} amounts to solving:
\begin{equation}
    \label{eq: parametric_umap_opt}
    \widehat{\phi} = \argmin_{\phi \in \Phi} C(P, Q_{\phi}),
\end{equation}
which can be accomplished using stochastic optimization. 

\section{Methodology}\label{sec: methodology}
In this work, we propose a novel variant of Mixup that joins manifold Mixup with parametric UMAP, which we call \emph{UMAP Mixup}. The idea is as follows:  we apply the Mixup operation in an intermediate layer of the classifier that is optimized to have similar topological structure to the original data using the UMAP loss as a regularizer.  

\subsection{Supervised Parametric UMAP}
We consider predictive models of the following form:
\begin{align}
    y = f_\theta(x) = g_{\theta_2}(h_{\theta_1}(x)),
\end{align}
where $h_{\theta_1}: \mathbb{R}^{d_x}\rightarrow \mathbb{R}^{d_z}$ is a neural network mapping from features to a $d_z$-dimensional embedding and $g_{\theta_2}:\mathbb{R}^{d_z}\rightarrow{\cal Y}$ is another neural network mapping the embeddings to predictions. The parameters $\theta=\{\theta_1, \theta_2\}$ are learned in a semi-supervised fashion using a loss $\ell(f_{\theta}(x), y)$ (supervised loss) and the UMAP loss as defined in \eqref{eq: umap_loss} (unsupervised loss). This is done by adding a regularizer to the risk:
\begin{equation}
    \label{eq: expected_risk_with_umap}
    R_{\rm reg}(\theta; \gamma) = \mathbb{E}[\ell(f_{\theta}(x), y)] + \gamma C(P, Q_{\theta_1}).
\end{equation}
where $Q_{\theta_1}$ is the parametric UMAP embedding graph as determined by the neural network $h_{\theta_1}$ and $\gamma\geq 0$ is a regularization parameter that controls the influence of the UMAP loss. The first term can be approximated using the empirical distribution of the data, just as it is done in ERM, while the second term can be approximated as by sampling edges of the data graph $P$. Specifically, for a given dataset ${\cal D}$, the \emph{supervised parametric UMAP loss function} is given by:
\begin{equation}
    \label{eq: supervised_parametric_umap}
    \widehat{R}_{\rm reg}(\theta; \gamma) = \left(\frac{1}{N}\sum_{i=1}^N\ell_{\theta}(f_\theta(x_i), y_i)\right) + \gamma C(P, Q_{\theta_1}),
\end{equation}
In practice, the parameters $\theta_1$ and $\theta_2$ can be learned jointly using stochastic optimization algorithms, such as stochastic gradient descent or Adam. In particular, each term in the loss in \eqref{eq: supervised_parametric_umap} is approximated by mini-batching, where the supervised component of the loss is approximated by mini-batching over the dataset ${\cal D}$ and the UMAP regularizer is approximated by mini-batching over the edges of the data graph $P$, as described in \cite{mcinnes2018umap} and \cite{sainburg2021parametric}. Please see Subsection \ref{ss: implementation} for more details on how we efficiently mini-batch in this work.




\subsection{UMAP Mixup}
Our proposed method UMAP Mixup is an extension of supervised parametric UMAP that  constructs augmentations by applying Mixup to a parametric UMAP embedding. During training, a single forward pass of our algorithm can be summarized in five steps:
\begin{enumerate}
    \item Generate a sample $(x, y)\sim{\cal D}$, where ${\cal D}$ denotes the empirical distribution of the data.
    \item Generate a neighboring sample $(x', y')$ based on the weighted data graph $P$. Note: the first two steps can be summarized into a single step by simply sampling a random edge from $P$.
    \item Generate a mixing ratio $\lambda\sim{\rm Beta}(\alpha, \alpha)$.
    \item Set $z=h_{\theta_1}(x)$ and $z'=h_{\theta_1}(x')$. Generate an interpolated embedding using the samples $(x, y)$ and $(x', y')$:
    \begin{equation}
        \tilde{z} = \lambda z + (1-\lambda)z'.
    \end{equation}
    \item Obtain the interpolated prediction as $\tilde{y}=g_{\theta_2}(\tilde{z})$.
    \end{enumerate}
The loss for the generated interpolated predictions is evaluated against an interpolation of the ground truth labels $y$ and $y'$:
\begin{equation}
    \label{eq: loss_umap_mixup}
    \ell_{\theta}^{{\rm mix}}(x, x', y, y') = \ell_\theta(\tilde{y}, \lambda y + (1-\lambda)y')
\end{equation}
In UMAP Mixup, the expected value of \eqref{eq: loss_umap_mixup}, augmented with the UMAP loss $C(P, Q_{\theta_1})$ is optimized in order to learn the model parameters:
\begin{equation}
    \label{eq: umap_mixup_params}
    \widehat{\theta}_{\rm UMAP}^{\rm mix} = \argmin_{\theta\in\Theta} \mathbb{E}[\ell_{\theta}^{{\rm mix}}(x, x', y, y')] + \gamma C(P, Q_{\theta_1}),
\end{equation}
where $\alpha$ and $\gamma$ are hyperparameters. In practice, these can be selected using cross-validation, where the choices vary from dataset to dataset. Empirically, we have found that choices of $\alpha=2$ and $\gamma=0.1$ lead to favorable results in practice for most datasets.


\subsection{Implementation Details} \label{ss: implementation}
Typical practice for creating batches consists of randomly shuffling the training dataset each epoch and iterating over subsets, effectively sampling without replacement. However, the cross entropy criterion in UMAP is defined over the set of edges of $P$. Our approach consists of creating mini-batches over the positive ($p_{i,j} > 0$) and negative ($p_{i,j} = 0$) edges, with the vertices that make up the batched positive edges then comprising the minibatch for the supervised loss.

Let $E^+ = \{e_{i,j}=(i, j)\}$  be the set of positive edges in a training epoch, where each edge $e_{i,j}$ is included in $E^+$ with probability $p_{i,j}$. For each positive edge $e_{i,j}$, we also associate $M$ negative edges $\{e_{i,j_{1}}, ..., e_{i,j_{M}}\}$ by sampling $j_1, ..., j_M$ uniformly from the dataset, giving the set of negative edges in a training epoch $E^- = \{e_{i,j_1}, ..., e_{i,j_M} : e_{i,j} \in E^+\}$. Each training epoch of UMAP Mixup therefore consists of iterating over subsets $E_b = E^+_b \cup E^-_b$, where $E^+_{b} \subset E^+$, $E^-_{b} \subset E^-$. This gives a batch UMAP loss of
\begin{align}
    \label{eq: umap_loss_batch}
    &C(P, Q_{\theta_1}) \approx \widehat{C}(P, Q_{\theta_1}, E_b)\\
    &=\frac{1}{|E_b|} \left( \sum_{e_{i,j} \in E^+_{b}} \hspace{-0.2cm}\log \left(\frac{p_{i, j}}{q_{i, j}}\right) + \hspace{-0.25cm}\sum_{e_{k,l} \in E^-_{b}} \hspace{-0.2cm}\log \left( \frac{1-p_{k, l}}{1-q_{k, l}}\right) \right)
\end{align}
The mini-batched supervised loss is then defined over the set of data points that occur as vertices of the positive edges in the batch: 
\begin{align}
    \label{eq: supervised_parametric_umap_batch}
    \mathbb{E}[\ell_{\theta}^{{\rm mix}}(x, x', y, y')] &\approx \widehat{\mathbb{E}}[\ell_{\theta}^{{\rm mix}}(x, x', y, y')]\\
    &=\frac{1}{|E^+_b|}\sum_{e_{i,j} \in E^+_{b}}\ell_{\theta}^{{\rm mix}}(x_i, x_j, y_i, y_j).
\end{align}

\begin{figure*}[t]
    \centering    \includegraphics[width=0.85\linewidth,trim={0 10.75cm 3cm 0},clip]{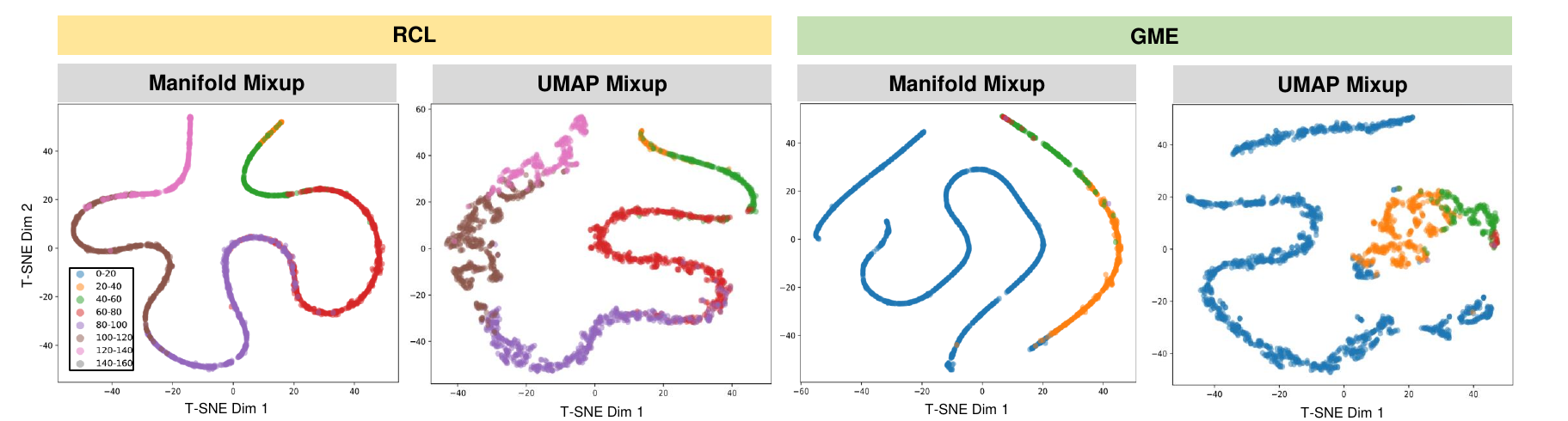}
    \caption{A visual comparison of resulting embeddings from both Manifold Mixup and UMAP Mixup regularizations on the RCL and GME datasets. Visualizations are obtained by applying T-SNE to the extracted features just before the output layer of each neural network.}
    \label{fig: projected_embeddings}
\end{figure*}

\vspace{-0.2cm}
\section{Experiments}
We validate UMAP Mixup on regression tasks covering two different data modalities: tabular data and time series data. All results are sumamrized in Table \ref{tab:results}, which measure performance using root-mean squared error (RMSE).

\subsection{Tabular Data}

We evaluate the performance of UMAP Mixup on a set of UCI regression benchmark datasets: \cite{Dua2019UCI} (a) Boston Housing dataset\footnote{\url{https://www.kaggle.com/datasets/schirmerchad/bostonhoustingmlnd}}, (b) Concrete compressive strength dataset\footnote{\url{https://archive.ics.uci.edu/ml/datasets/concrete+compressive+strength}} \cite{yeh1998modeling}, and (c) Yacht hydrodynamics dataset\footnote{\url{https://archive.ics.uci.edu/ml/datasets/Yacht+Hydrodynamics}}.
We use the experimental setup used in \cite{el-laham2023deep}, with each dataset split into 20 train-test folds. For each method, we utilize a $(100, 50)$  2-layer feedforward neural network. We train each method using the Adam optimizer, where we select learning rates and batch sizes according to values utilized in \cite{sarawgi2020have}. For each baseline, we use cross-validation to select all other hyperparameters. The results in Table \ref{tab:results} indicate that for the Concrete and Yacht datasets, UMAP Mixup is able to get significant reduction in RMSE as compared to the baselines.

\begin{table}[t]
    \centering
    \caption{Experiment results.}
    \label{tab:results}
    \resizebox{\columnwidth}{!}{
\begin{tabular}{lllll}
\toprule
                   Dataset &    ERM &  Mixup & Manifold Mixup & UMAP Mixup \\
\midrule
      \multicolumn{1}{l}{\textit{Tabular (Regression)}} & & &                      &       \\
\midrule
            Boston Housing & 3.14 $\pm$  0.67 &  \bf 3.01 $\pm$  0.71 &  3.10 $\pm$  0.76 &  3.27 $\pm$  0.66 \\
                  Concrete & 5.11 $\pm$  0.59 &      5.92 $\pm$  0.55 &  5.08 $\pm$  0.62 &  \bf 4.83 $\pm$  0.79 \\
                     Yacht & 0.91 $\pm$  0.34 &      4.19 $\pm$  0.63 &  0.80 $\pm$  0.24 &  \bf 0.71 $\pm$  0.21 \\
\midrule
      \multicolumn{1}{l}{\textit{Time series (Regression)}} &    &  &                   &       \\
\midrule
    GOOG    & 2.47 $\pm$ 0.05  &  2.47 $\pm$ 0.03   &	2.50 $\pm$ 0.03	& \bf 2.43 $\pm$ 0.04        \\
    RCL     & 4.74 $\pm$ 0.69  &  4.07 $\pm$ 0.43	&   4.30 $\pm$ 0.60	& \bf 3.13 $\pm$ 0.61         \\
    GME     & 3.66 $\pm$ 0.33  &  2.77 $\pm$ 0.49	&   3.83 $\pm$ 0.47	& \bf 2.73 $\pm$ 0.37         \\
\bottomrule
\end{tabular}
 }
\end{table}


\subsection{Time Series Data}

For the time series data, we focus on the task of one-step-ahead forecasting for financial time series. We use historical daily price data from Yahoo finance 
and follow a similar  experimental set up as \cite{el-laham2023deep}, using a long short-term memory (LSTM) network \cite{LSTMhochreiter}. The input to the LSTM corresponds to the closing price of a particular stock over the past 60 trading days and the target output is the next trading day’s closing price. We evaluate each method on three different datasets:
\begin{itemize}
    \item {\bf GOOG - stable market regime:} We use training data from the Google (GOOG) stock from the period of Jan 2015 - July 2022 and test on GOOG stock data from the period of August 2022 - September 2023. 
    \item {\bf RCL - market shock regime:} We use training data from the Royal Caribbean (RCL) stock from the period of Jan 2015 - June 2020 and test on RCL stock data from the period of July 2020 - September 2023.  This choice of dataset is motivated by the COVID shock that heavily affected the Royal Caribbean stock when the global pandemic caused widespread closures in the spring of 2020,
    \item {\bf GME - high volatility regime:} We use the training data from the Gamestop (GME) stock to include the short squeeze period of January 2015 - Jan 2022 and test on GME stock data following that period. 
\end{itemize}

The results are shown in Table \ref{tab:results}, where UMAP Mixup outperforms all baselines. In particular, Manifold Mixup shows worst performance than the other methods, indicating that interpolation in the embedding space without further regularization does not necessarily improve generalization. Figure \ref{fig: projected_embeddings} shows a qualitative comparison of the embeddings of Manifold Mixup and our method, Manifold UMAP for the RCL and GME datasets by using T-SNE to project the extracted features on the last hidden layer of each neural network. We can observe that Manifold Mixup shows a very compact embedding whilst the UMAP Mixup embeddings show more variability and therefore, interpolation between them would yield more diverse samples. This is particularly relevant in these datasets that suffer from distributional shifts such as a market shock regime in the case of RCL due to COVID, and the high volatility regime of GME during the "bubble" period in early 2021.


\section{Conclusion}
\label{sec:conclusion}
This work proposes a novel variant of Mixup regularization called UMAP Mixup. UMAP Mixup hybridizes a nonlinear dimensionality reduction technique called UMAP with Mixup regularization.  Unlike other variants of Mixup, such as Manifold Mixup, UMAP Mixup synthesizes augmentations in a hidden layer of neural network that is regularized to have nice topological properties: mainly, that the optimized manifold in UMAP Mixup better preserves global and local structure as compared to its counterparts. Our results on empirical evaluations across three different data modalities (tabular, image, and time series data) show that UMAP Mixup performs well in comparison to multiple state-of-the-art baselines. 

\section{Acknowledgements}
This paper was prepared for informational purposes by the Artificial Intelligence Research group of JPMorgan Chase \& Co. and its affiliates (``JP Morgan''), and is not a product of the Research Department of JP Morgan. JP Morgan makes no representation and warranty whatsoever and disclaims all liability, for the completeness, accuracy or reliability of the information contained herein. This document is not intended as investment research or investment advice, or a recommendation, offer or solicitation for the purchase or sale of any security, financial instrument, financial product or service, or to be used in any way for evaluating the merits of participating in any transaction, and shall not constitute a solicitation under any jurisdiction or to any person, if such solicitation under such jurisdiction or to such person would be unlawful.

\bibliographystyle{IEEEtran}
\bibliography{ref}

\end{document}